\documentclass[runningheads]{llncs}
\usepackage{graphicx}
\usepackage{amsmath,amssymb} % define this before the line numbering.
\usepackage{color}
\usepackage{amsmath,amssymb} % define this before the line numbering.

\usepackage{amsfonts}
\usepackage{tabularx}
\usepackage{booktabs}
\usepackage{subcaption}
\usepackage{array}

\usepackage[breaklinks]{hyperref}

\newcommand{\fig}[1]{Figure~\ref{fig:#1}}

\newcommand{\tab}[1]{Table~\ref{tab:#1}}

\newcommand{\eq}[1]{(\ref{eq:#1})}

\newcommand{\x}{\mathbf{x}}
\newcommand{\y}{\mathbf{y}}

\makeatletter
\newcommand{\printfnsymbol}[1]{%
  \textsuperscript{\@fnsymbol{#1}}%
}
\makeatother

\begin{document}
\title{Image Manipulation with\\ Perceptual Discriminators} 
% Replace with your title

\titlerunning{Image Manipulation with Perceptual Discriminators}
% Replace with a meaningful short version of your title
%
\author{Diana Sungatullina\thanks{indicates equal contribution} \and
Egor Zakharov\printfnsymbol{1} \and
Dmitry Ulyanov \and 
Victor Lempitsky}
%
%Please write out author names in full in the paper, i.e. full given and family names. 
%If any authors have names that can be parsed into FirstName LastName in multiple ways, please include the correct parsing, in a comment to the volume editors:
%\index{Lastnames, Firstnames}
%(Do not uncomment it, because you may introduce extra index items if you do that, we will use scripts for introducing index entries...)
\authorrunning{D. Sungatullina*, E. Zakharov*,  D. Ulyanov, and V. Lempitsky}
% Replace with shorter version of the author list. If there are more authors than fits a line, please use A. Author et al.
%

\institute{Skolkovo Institute of Science and Technology, Moscow, Russia \\ 
\email{\{d.sungatullina, egor.zakharov, dmitry.ulyanov, lempitsky\}@skoltech.ru}}
\maketitle              % typeset the header of the contribution
\begin{abstract}
Systems that perform image manipulation using deep convolutional networks have achieved remarkable realism. Perceptual losses and losses based on adversarial discriminators are the two main classes of learning objectives behind these advances. In this work, we show how these two ideas can be combined in a principled and non-additive manner for unaligned image translation tasks. This is accomplished through a special architecture of the discriminator network inside generative adversarial learning framework. The new architecture, that we call a \textit{perceptual discriminator}, embeds the convolutional parts of a pre-trained deep classification network inside the discriminator network. The resulting architecture can be trained on unaligned image datasets, while benefiting from the robustness and efficiency of perceptual losses. We demonstrate the merits of the new architecture in a series of qualitative and quantitative comparisons with baseline approaches and state-of-the-art frameworks for unaligned image translation.

\end{abstract}
\section{Introduction} \label{section_intro}

Generative convolutional neural networks have achieved remarkable success in image manipulation tasks both due to their ability to train on large amount of data \cite{Jain09,Kim16,Dosovitskiy15} and due to natural image priors associated with such architectures~\cite{Ulyanov18}.
Recently, the ability to train image manipulation ConvNets has been shown in the \textit{unaligned} training scenario \cite{Zhu17cycle,Zhu17bicycle,Benaim17}, where the training is based on sets of images annotated with the presence/absence of a certain attribute, rather than based on \textit{aligned} datasets containing \{input,output\} image pairs. The ability to train from unaligned data provides considerable flexibility in dataset collection and in learning new manipulation effects, yet poses additional algorithmic challenges.

Generally, the realism of the deep image manipulation methods is known to depend strongly on the choice of the loss functions that are used to train generative ConvNets. In particular, simplistic pixelwise losses (e.g.\ the squared distance loss) are known to limit the realism and are also non-trivial to apply in the unaligned training scenario. The rapid improvement of realism of deep image generation and processing is thus associated with two classes of loss functions that go beyond pixel-wise losses. The first group (so-called \textit{perceptual losses}) is based on matching activations inside pre-trained deep convolutional networks (the VGG architecture trained for ILSVRC image classification is by far the most popular choice \cite{Simonyan14}). The second group consists of \textit{adversarial losses}, where the loss function is defined implicitly using a separate \textit{discriminator} network that is trained adversarially in parallel with the main generative network.

The two groups (perceptual losses and adversarial losses) are known to have largely complementary strengths and weaknesses. Thus, perceptual losses are easy to incorporate and are easy to scale to high-resolution images; however, their use in unaligned training scenario is difficult, as these loss terms require a concrete target image to match the activations to. Adversarial losses have the potential to achieve higher realism and can be used naturally in the unaligned scenarios, yet adversarial training is known to be hard to set up properly, often suffers from mode collapse, and is hard to scale to high-resolution images. Combining perceptual and adversarial losses in an additive way has been popular \cite{Dosovitskiy16,Wang17,Ledig17,Sajjadi17}. Thus, a generative ConvNet can be trained by minimizing a linear combination of an adversarial and a perceptual (and potentially some other) losses. Yet such additive combination includes not only strengths but also weaknesses of the two approaches. In particular, the use of a perceptual loss still incurs the use of aligned datasets for training.

In this work we present an architecture for realistic image manipulation, which combines perceptual and adversarial losses in a natural \textit{non-additive} way. Importantly, the architecture keeps the ability of adversarial losses to train on unaligned datasets, while also benefits from the stability of perceptual losses. Our idea is very simple and concerned with the particular design of the discriminator network for adversarial training. The design encapsulates a pretrained classification network as the initial part of the discriminator. During adversarial training, the generator network is  effectively learned to match the activations inside several layers of this reference network, just like the perceptual losses do. We show that the incorporation of the pretrained network into the discriminator stabilizes the training and scales well to higher resolution images, as is common with perceptual losses. At the same time, the use of adversarial training allows to avoid the need for aligned training data.

Generally, we have found that the suggested architecture can be trained with little tuning to impose complex image manipulations, such as adding to and removing smile from human faces, face ageing and rejuvenation, gender change, hair style change, etc. In the experiments, we show that our architecture can be used to perform complex manipulations at medium and high resolutions, and compare the proposed architecture with several adversarial learning-based baselines and recent methods for learning-based image manipulation.

\section{Related work} \label{section_related}

\subsubsection{Generative ConvNets.} Our approach is related to a rapidly growing body of works on ConvNets for image generation and editing. Some of the earlier important papers on ConvNet image generation \cite{Dosovitskiy15} and image processing \cite{Jain09,Dong14,Kim16} used per-pixel loss functions and fully supervised setting, so that at test time the target image is known for each input. While this demonstrated the capability of ConvNets to generate realistic images, the proposed systems all had to be trained on aligned datasets and the amount of high-frequency details in the output images was limited due to deficiencies of pixel-wise loss functions.

\subsubsection{Perceptual Losses.} The work of Mahendran and Vedaldi~\cite{Mahendran15} has demonstrated that the activations invoked by an image within a pre-trained convolutional network can be used to recover the original image. Gatys et al.~\cite{Gatys16} showed that such activations can serve as content descriptors or texture descriptors of the input image, while Dosovitsky and Brox~\cite{Dosovitskiy16}, Ulyanov~et~al.~\cite{Ulyanov16}, Johnson~et~al.~\cite{Johnson16} have shown that the mismatches between the produced and the target activations can be used as so-called \textit{perceptual losses} for a generative ConvNet. The recent work of \cite{Chen17} pushed the spatial resolution and the realism of images produced by a feed-forward ConvNet with perceptual losses to megapixel resolution.  Generally, in all the above-mentioned works \cite{Chen17,Ulyanov16,Johnson16,Dosovitskiy16}, the perceptual loss is applied in a fully supervised manner as for each training example the specific target deep activations (or the Gram matrix thereof) are given explicitly. Finally, \cite{Upchurch17} proposed a method that manipulates carefully aligned face images at high resolution by compositing desired activations of a deep pretrained network and finding an image that matches such activations using the non-feedforward optimization process similar to \cite{Mahendran15,Gatys16}.

\subsubsection{Adversarial Training.} The most impressive results of generative ConvNets were obtained within generative adversarial networks (GANs) framework proposed originally by Goodfellow~et~al.~\cite{Goodfellow14}. The idea of adversarial training is to implement the loss function as a separate trainable network (the \textit{discriminator}), which is trained in parallel and in adversarial way with the generative ConvNet (the \textit{generator}). Multiple follow-up works including~\cite{Radford15,Salimans16,Arjovsky17,Karras17} investigated the choice of convolutional architectures for the generator and for the discriminator. Achieving reliable and robust convergence of generator-discriminator pairs remains challenging \cite{Goodfellow17,ganhacks,Lucic17}, and in particular requires considerably more efforts than training with perceptual loss functions.

\subsubsection{Unaligned Adversarial Training.} While a lot of the original interest to GANs was associated with unconditional image generation, recently the emphasis has shifted to the conditional image synthesis. Most relevant to our work are adversarially-trained networks that perform image translation, i.e.~generate output images conditioned on input images. While initial methods used aligned datasets for training \cite{Zhang16,Isola17}, recently some impressive results have been obtained using unaligned training data, where only empirical distributions of the input and the output images are provided \cite{Zhu17cycle,Benaim17,Zhu17bicycle}. For face image manipulation, systems using adversarial training on unaligned data have been proposed in \cite{Brock16,Choi18}. While we also make an emphasis on face manipulation, our contribution is orthogonal to \cite{Brock16,Choi18} as perceptual discriminators can be introduced into their systems. 

\subsubsection{Combining Perceptual and Adversarial Losses.} A growing number of works \cite{Dosovitskiy16,Ledig17,Wang17} use the combination of perceptual and adversarial loss functions to accomplish more stable training and to achieve convincing image manipulation at high resolution. Most recently, \cite{Sajjadi17} showed that augmenting perceptual loss with the adversarial loss improves over the baseline system \cite{Chen17} (that has already achieved very impressive results) in the task of megapixel-sized conditional image synthesis. Invariably, the combination of perceptual and adversarial losses is performed in an additive manner, i.e.\ the two loss functions are weighted and added to each other (and potentially to some other terms). While such additive combination is simple and often very efficient, it limits learning to the aligned scenario, as perceptual terms still require to specify target activations for each training example. In this work, we propose a natural non-additive combination of perceptual losses and adversarial training that avoids the need for aligned data during training. %Finally, we note that adversarial learning and perceptual features have been combined in a non-linear way in \cite{Nguyen17}, however in general their system is very different from ours (solves a different problem) and the manner  

\section{Perceptual discriminators} \label{section_method}

\begin{figure}[t]
    % \hspace*{-2cm}
    \centering
    %\def\svgwidth{\columnwidth}
    %\hspace*{-1.5cm}\input{per}
    \includegraphics[width=0.9\textwidth]{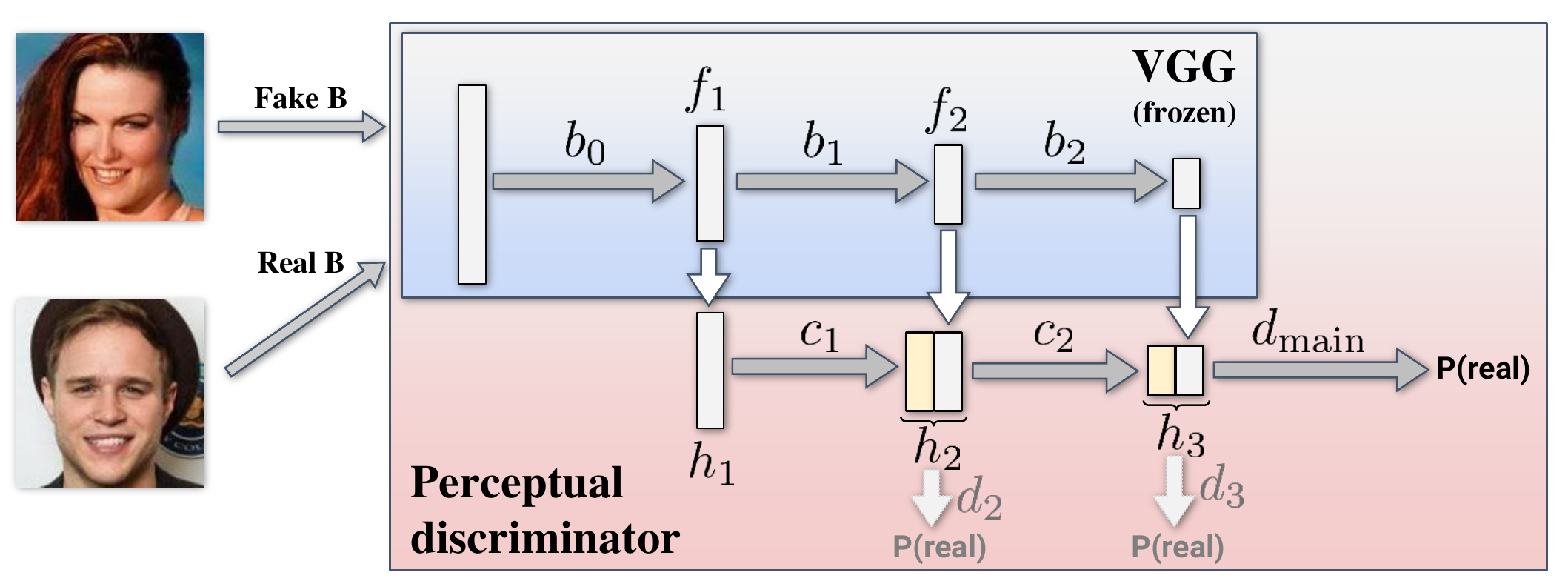}
    \caption{The perceptual discriminator is composed of a pre-trained image classification network (such as VGG), split into blocks $b_i$. The parameters of those blocks are not changed during training, thus the discriminator retains access to so-called perceptual features. The outputs of these blocks are processed using learnable blocks of convolutional operations $c_i$ and the outputs of those are used to predict the probability of an image being real or manipulated (the simpler version uses a single discriminator $d_\text{main}$, while additional path discriminators are used in the full version).}
    \label{fig:model}
\end{figure}

\subsection{Background and motivation}

Generative adversarial networks have shown impressive results in photorealistic image synthesis. The model includes a generative network $G$, that is trained to match the target distribution $p_\text{target}(\y)$ in the data space $\mathcal Y$, and a discriminator network $D$ that is trained to distinguish whether the input is real or generated by $G$. In the simplest form, the two networks optimize the policy function $V(D, G)$:
\begin{equation}
    \min_{G} \max_{D} V(D, G) = \mathbb{E}_{\y \sim p_\text{target}(\y)} \log D(\y) + \mathbb{E}_{\x \sim  p_\text{source}(\x)} [\log (1 - D(G(\x))] ,
\label{eq:gan}
\end{equation}
In \eq{gan}, the source distribution $p_\text{source}(\x)$  may correspond to a simple parametric distribution in a latent space such as the unit Gaussian, so that after training unconditional samples from the learned approximation to $p_\text{target}(\y)$ can be drawn. Alternatively, $p_\text{source}(\x)$ may correspond to another empirical distribution in the image space $\mathcal X$. In this case, the generator learns to \textit{translate} images from $\mathcal X$ to $\mathcal Y$, or to \textit{manipulate} images in the space $\mathcal X$ (when it coincides with $\mathcal Y$). Although our contribution (perceptual discriminators) is applicable to both unconditional synthesis and image manipulation/translation, we focus our evaluation on the latter scenario. For the low resolution datasets, we use the standard non-saturating GAN modification, where the generator maximizes the log-likelihood of the discriminator instead of minimizing the objective \eq{gan} \cite{Goodfellow14}. For high-resolution images, following CycleGAN~\cite{Zhu17cycle}, we use the LSGAN formulation~\cite{MaoLXLW16}.

% Converging to good equilibria for any of the proposed GAN games is known to be hard \cite{Goodfellow17,ganhacks,Lucic17}. In general, the performance of the trained generator network crucially depend on the architecture of the discriminator network. This has been observed empirically as improvements in the generator architecture has been reported to improve the quality and the diversity of the generator output \cite{Radford15}. Furthermore, \cite{Arora17} has suggested bounds that draw a connection between the support of the generator output in $\mathcal Y$ on one hand, and the number of parameters in the discriminator on the other. Generally, the discriminator network needs to learn to extract meaningful statistics, which are good for matching the target distribution $p_\text{target}$. The typical failure mode of GAN training is when the discriminator does not manage to learn such statistics before being ``overpowered'' by the generator. 

Converging to good equilibria for any of the proposed GAN games is known to be hard \cite{Goodfellow17,ganhacks,Lucic17}. In general, the performance of the trained generator network crucially depends on the architecture of the discriminator network, that needs to learn meaningful statistics, which are good for matching the target distribution $p_\text{target}$. The typical failure mode of GAN training is when the discriminator does not manage to learn such statistics before being ``overpowered'' by the generator. 

% where $z$ is an input noise and $x$ is a data sample. However, in case of image editing, when one alters only chosen properties of an image without modification the whole scene, more suitable to match distribution of input images $x$ without an attribute and distribution of images $y$ with the attribute, so
% \begin{equation}
%     \min_{G} \max_{D} V(D, G) = \mathbb{E}_{y \sim p_{t}(y)} \log D(y) + \mathbb{E}_{x \sim p_{s}(x)} [\log (1 - D(G(x)] ,
% \label{eq:gan-ours}
% \end{equation}
% where $p_{s}(x)$ and $p_{t}(y)$ are data distributions of source and target domains, respectively. In practise, folks solve the equivalent $\max [D(G(x))]$ instead of $\min \log [1 - D(G(x)]$ in order to stabilize the gradients~\cite{Goodfellow14} and play non-zero sum adversarial game.

\subsection{Perceptual Discriminator Architecture}

Multiple approaches have suggested to use activations invoked by an image $\y$ inside a deep pre-trained classification network $F(\y)$ as  statistics for such tasks as retrieval \cite{Babenko14} or few-shot classification \cite{Razavian14}. Mahendran and Vedaldi~\cite{Mahendran15} have shown that activations computed after the convolutional part of such network retain most of the information about the input $\y$, i.e.\ are essentially invertable. Subsequent works such as \cite{Gatys16,Ulyanov16,Johnson16,Dosovitskiy16} all used such ``perceptual'' statistics to match low-level details such as texture content, certain image resolution, or particular artistic style.

Following this line of work, we suggest to base the GAN discriminator $D(\y)$ on the perceptual statistics computed by the reference network $F$ on the input image $\y$, which can be either real (coming from $p_\text{target}$) or fake (produced by the generator). Our motivation is that a discriminator that uses perceptual features has a better chance to learn good statistics than a discriminator initialized to a random network. For simplicity, we assume that the network $F$ has a chain structure, e.g.\ $F$ can be the VGGNet of \cite{Simonyan14}.

Consider the subsequent blocks of the convolutional part of the reference network $F$, and denote them as $b_0,b_1,\dots,b_{K-1}$. Each block may include one or more convolutional layers interleaved with non-linearities and pooling operations. Then, the perceptual statistics $\{f_1(\y), \dots, f_K(\y)\}$ are computed as:
\begin{eqnarray}
f_1(\y) &=& b_0(\y)\\
f_i(\y) &=& b_{i-1}(f_{i-1}(\y)), \quad i = 2,\dots,K\,,
\end{eqnarray}
so that each $f_i(\y)$ is a stack of convolutional maps of the spatial dimensions $W_i \times W_i$. The dimension $W_i$ is determined by the preceeding size $W_{i-1}$ as well as by the presence of strides and pooling operations inside $b_i$. In our experiments we use features from consecutive blocks, i.e. $W_i = W_{i-1} / 2$.
% NEW

The overall structure of our discriminator is shown in \fig{model}. The key novelty of our discriminator is the in-built perceptual statistics $f_i$ (top of the image), which are known to be good at assessing image realism \cite{Gatys16,Johnson16,Upchurch17}. During the backpropagation, the gradients to the generator flow through the perceptual statistics extractors $b_i$, but the parameters of $b_i$ are frozen and inherited from the network pretrained for large-scale classification. This stabilizes the training, and ensures that at each moment of time the discriminator has access to ``good'' features, and therefore cannot be overpowered by the generator easily. 

In more detail, the proposed discriminator architecture combines together perceptual statistics using the following computations:
\begin{eqnarray}
h_1(\y) &=& f_1(\y)\\
h_i(\y) &=& \texttt{stack}\left[ c_{i-1}(h_{i-1}(\y),\phi_{i-1})\,,\, f_i(\y) \right], \quad i = 2,\dots,K\,, \label{eq:midlayer}
\end{eqnarray}
where \texttt{stack} denotes stacking operation, and the convolutional blocks $c_j$ with learnable parameters $\phi_j$ (for $j = 1,\dots,K-1$) are composed of convolutions, leaky ReLU nonlinearities, and average pooling operations. Each of the $c_j$ blocks thus transforms map stacks of the spatial size $W_j \times W_j$ to map stacks of the spatial size  $W_{j+1} \times W_{j+1}$. Thus, the strides and pooling operations inside $c_j$ match the strides and/or pooling operations inside $b_j$.

Using a series of convolutional and fully-connected layers with learnable parameters $\psi_\text{main}$ applied to the representation $h_K(\y)$, the discriminator outputs the probability $d_\text{main}$ of the whole image $\y$ being real. For low- to medium- resolution images we perform experiments using only this probability. For high-resolution, we found that additional outputs from the discriminator resulted in better outcomes. Using the ``patch discriminator'' idea \cite{Isola17,Zhu17cycle}, to several feature representations $h_j$ we apply a convolution+LeakyReLU block $d_j$ with learnable parameters $\psi_j$ that outputs probabilities $d_{j,p}$ at every spatial locations $p$. We then replace the regular log probability $\log D(\y) \equiv \log d_\text{main}$ of an image being real with:
\begin{equation}
    \log D(\y) = \log d_\text{main}(\y) + \sum_j \sum_{p \in \text{Grid}(W_j\times W_j)} \log d_{j,p}(\y)\,
\end{equation}
Note, that this makes our discriminator ``multi-scale'', since spatial resolution $W_j$ varies for different $j$. The idea of multiple classifiers inside the discriminator have also been proposed recently  in \cite{Wang17,Iizuka17}. Unlike \cite{Wang17,Iizuka17} where these classifiers are disjoint, in our architecture all such classifiers are different branches of the same network that has perceptual features underneath.

During training, the parameters of the $c$ blocks inside the feature network $F$ remain fixed, while the parameters $\phi_i$ of feature extractors $c_i$ and the parameters $\psi_i$ of the discriminators $d_i$ are updated during the adversarial learning, which forces the ``perceptual'' alignment between the output of the generator and $p_\text{target}$. Thus,  wrapping perceptual loss terms into additional layers $c_i$ and $d_i$ and putting them together into the adversarial discriminator allows us to use such perceptual terms in the unaligned training scenario. Such unaligned training was, in general, not possible with the ``traditional'' perceptual losses. 

\subsection{Architecture Details} \label{sect:subsection_archdetails}

\subsubsection{Reference Network.} \label{sect:reference_network} Following multiple previous works~\cite{Gatys16,Ulyanov16,Johnson16}, we consider the so-called \textit{VGG network} from \cite{Simonyan14} trained on ILSVRC2012~\cite{RussakovskyDSKSMHKKBBF14} as the reference network $F$. In particular, we pick the VGG-19 variant, to which we simply refer to as VGG. While the perceptual features from VGG already work well, the original VGG architecture can be further improved. Radford~et.~al~\cite{Radford15} reported that as far as leaky ReLU avoids sparse gradients, replacing ReLUs with leaky ReLUs \cite{He15b} in the discriminator stabilizes the training process of GANs. For the same reasons, changing max pooling layers to average pooling removes unwanted sparseness in the backpropagated gradients. Following these observations, we construct the \textit{VGG*} network, which is particularly suitable for the adversarial game. We thus took the VGG-19 network pretrained on ILSVRC dataset, replaced all max pooling layers by average poolings, ReLU nonlinearities by leaky ReLUs with a negative slope $0.2$ and then trained on the ILSVRC dataset for further two days. We compare the variants of our approach based on VGG and VGG* features below.

\subsubsection{Generator Architecture.} For the image manipulation experiments, we used transformer network proposed by Johnson et al.~\cite{Johnson16}. It consists of $M$ convolutional layers with stride size $2$, $N$ residual blocks~\cite{He15} and $M$ 
% upsample convolution layers to avoid the checkerboard effect~\cite{Odena16} with the scale factor $2\times$. 
upsampling layers, each one increases resolution by a factor of $2$. We set $M$ and $N$ in a way that allows outputs of the last residual block to have large enough receptive field, but at the same time for generator and discriminator to have similar number of parameters. We provide detailed descriptions of architectures in \cite{supmat}.

\subsubsection{Stabilizing the Generator.} We have also used two additional methods to improve the generator learning and to prevent its collapse. First, we have added the \textit{identity loss}~\cite{TaigmanPW16,Zhu17cycle} that ensures that the generator does not change the input, when it comes from the $p_\text{target}$. Thus, the following term is added to the maximization objective of the generator:
\begin{equation}\label{eq:identity_loss}
    J^G_\text{id} = -\lambda_\text{id} \, \mathbb{E}_{\y \sim p_\text{target}}  \lambda \bigl\| \y - G(\y) \bigr\|_{L_1} ,
\end{equation}
where $\lambda_\text{id}$ is a meta-parameter that controls the contribution of the weight, and $\|\cdot\|_{L_1}$ denotes pixel-wise L1-metric.

To achieve the best results for the hardest translation tasks, we have found the cycle idea from the CycleGAN \cite{Zhu17cycle} needed. We thus train two generators $G_{\x \to \y}$ and $G_{\y \to \x}$ operating in opposite directions in parallel (and jointly with two discriminators), while adding reciprocity terms ensuring that mappings $G_{\x \to \y}\circ G_{\y \to \x}$ and $G_{\y \to \x}\circ G_{\x \to \y}$ are close to identity mappings.

Moreover, we notice that usage of external features as inputs for the discriminator leads to fast convergence of the discriminator loss to zero. Even though this is expected, since our method essentially corresponds to pretraining of the discriminator, this behavior is one of the GAN failure cases \cite{ganhacks} and on practice leads to bad results in harder tasks. Therefore we find pretraining of the generator to be required for increased stability. For image translation task we pretrain generator as autoencoder. Moreover, the necessity to pretrain the generator makes our approach fail to operate in DCGAN setting with unconditional generator.

After an additional stabilization through the pretraining and the identity and/or cycle losses, the generator becomes less prone to collapse. Overall, in the resulting approach it is neither easy for the discriminator to overpower the generator (this is prevented by the identity and/or cycle losses), nor is it easy for the generator to overpower the discriminator (as the latter always has access to perceptual features, which are good at judging the realism of the output).

\section{Experiments} \label{section_results}

The goal of the experimental validation is two-fold. The primary goal is to validate the effect of perceptual discriminators as compared to baseline architectures which use traditional discriminators that do not have access to perceptual features. The secondary goal is to validate the ability of our full system based on perceptual discriminators to handle harder image translation/manipulation task with higher resolution and with less data. Extensive additional results are available on our project page~\cite{supmat}.
We perform the bulk of our experiments on CelebA dataset~\cite{liu2015faceattributes}, due to its large size, popularity and the availability of the attribute annotations (the dataset comprises over $200$k of roughly-aligned images with $40$ binary attributes; we use $160 \times 160$ central crops of the images). As harder image translation task, we use CelebA-HQ~\cite{Karras17} dataset, which consists of high resolution versions of images from CelebA and is smaller in size. Lastly, we evaluate our model on problems with non-face datasets like apples to oranges and photo to Monet texture transfer tasks.

Experiments were carried out on NVIDIA DGX-2 server.

\subsubsection{Qualitative Comparison on CelebA.} Even though our contribution is orthogonal to a particular GAN-based image translation method, we chose one of them, provided modifications we proposed and compared it with the following important baselines in an attribute manipulation task:

\begin{figure}[t]
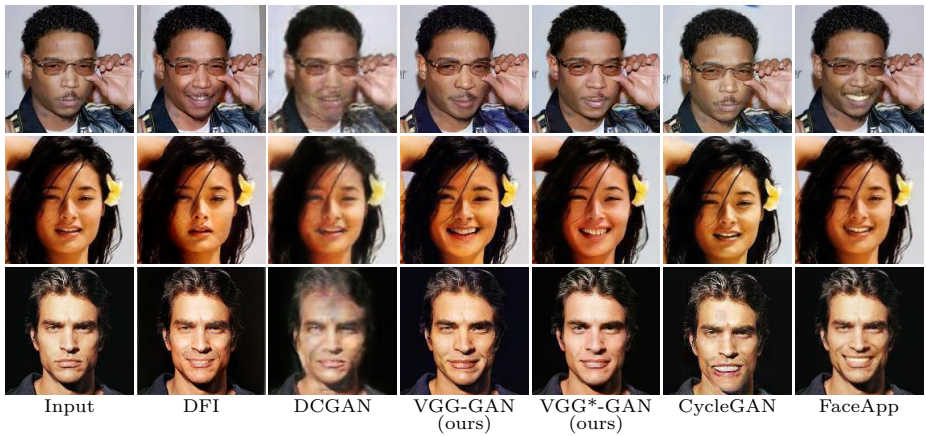

    \def\pth{images/Figure2/}
    \def\wid{1.70cm}
    \def\linefigure[#1]{\includegraphics[width=\wid]{\pth#1-input.png}&
    \includegraphics[width=\wid]{\pth#1-dfi.png}&
    \includegraphics[width=\wid]{\pth#1-dcgan.png}&
    \includegraphics[width=\wid]{\pth#1-vgg-plain.png}&
    \includegraphics[width=\wid]{\pth#1-vgg-smart.png}&
    \includegraphics[width=\wid]{\pth#1-cyclegan.png}&
    \includegraphics[width=\wid]{\pth#1-faceapp.png}\\}

    \centering
    \begin{scriptsize}
    \setlength{\tabcolsep}{0.7pt}
    \renewcommand{\arraystretch}{0.5}
    \begin{tabular}{@{}ccccccc@{}}
    \linefigure[0]
    \linefigure[1]
    \linefigure[2]
    Input& DFI&  DCGAN& \begin{tabular}[t]{@{}c@{}}VGG-GAN\\(ours)\end{tabular}&  \begin{tabular}[t]{@{}c@{}}VGG*-GAN\\(ours)\end{tabular}& CycleGAN& FaceApp
    \end{tabular}
    \end{scriptsize}
    \caption{Qualitative comparison of the proposed systems as well as baselines for neutral$\to$smile image manipulation. As baselines, we show the results of DFI (perceptual features, no adversarial training) and DCGAN (same generator, no perceptual features in the discriminator). Systems with perceptual discriminators output more plausible manipulations.}
    \label{fig:baselines-lr}
\end{figure}

\begin{itemize}
\item \textit{DCGAN}~\cite{Radford15}: in this baseline GAN system we used image translation model with generator and discriminator trained only with adversarial loss.

\item \textit{CycleGAN}~\cite{Zhu17cycle}: this GAN-based method learns two reciprocal transforms in parallel with two discriminators in two domains. We have used the authors' code (PyTorch version). 
% The architecture of generators is therefore slightly different from the other GAN-based systems.

\item \textit{DFI}~\cite{Upchurch17}: to transform an image, this approach first determines target VGG feature representation by adding the feature vector corresponding to input image and the shift vector calculated using nearest neighbours in both domains. Then the resulting image is produced using optimization-based feature inversion as in \cite{Mahendran15}. We have used the authors' code.

\item \textit{FaceApp}~\cite{FaceApp}: is a very popular closed-source app that is known for the quality of its filters (transforms), although the exact algorithmic details are unknown.
\end{itemize}

Our model is represented by two basic variants.

\begin{itemize}

    \item \textit{VGG-GAN}: we use DCGAN as our base model. The discriminator has a single classifier and no generator pretraining or regularization is applied, other than identity loss mentioned in the previous section.
    
    \item \textit{VGG*-GAN}: same as the previous model, but we use a finetuned VGG network variant with dense gradients.

\end{itemize}

The comparison with state-of-the-art image transformation systems is performed to verify the competitiveness of the proposed architecture (\fig{baselines-lr}). In general, we observe that VGG*-GAN and VGG-GAN models consistently outperformed DCGAN variant, achieving higher effective resolution and obtaining more plausible high-frequency details in the resulting images. While a more complex CycleGAN system is also capable of generating crisp images, we found that the synthesized smile often does not look plausible and does not match the face. DFI turns out to be successful in attribute manipulation, yet often produces undesirable artifacts, while FaceApp shows photorealistic results, but with low attribute diversity. Here we also evaluate the contribution of dense gradients idea for VGG encoder and find it providing minor quality improvements.

\begin{table}[t]   
    \caption{Quantitative comparison: (a) Photorealism user study. We show the fraction of times each method has been chosen as ``the best'' among all in terms of photorealism and identity preservation (the higher the better). (b) C2ST results (cross-entropy, the higher the better). (c) Log-loss of classifier trained on real data for each class (the lower the better). See main text for details.\\}
    
    \begin{subtable}[t]{\textwidth}
        \centering
        \begin{tabular}{
                c | 
                *{2}{>{\centering\arraybackslash}p{1.15cm}} | 
                *{3}{>{\centering\arraybackslash}p{1.15cm}} | 
                *{3}{>{\centering\arraybackslash}p{1.15cm}}
            }
            & \multicolumn{2}{c|}{(a) User study} 
            & \multicolumn{3}{c|}{(b) C2ST, $\times 10^{-2}$}
            & \multicolumn{3}{c}{(c) Classification loss} \\
            \cline{2-9}
                                       & Smile & Age  & Smile   & Gender  & Hair color & Smile & Gender & Hair color \\ \hline
            DFI~\cite{Upchurch17}      & 0.16  & 0.4  & $<0.1$  & $<0.01$ & $<0.01$    & 1.3   & 0.5    & 1.14       \\
            FaceApp~\cite{FaceApp}     & 0.45  & 0.41 & --      & --      & --         & --    & --     & --         \\
            DCGAN~\cite{Radford15}     & --    & --   & 0.6     & 0.03    & 0.06       & 0.6   & 1.5    & 2.33       \\
            CycleGAN~\cite{Zhu17cycle} & 0.03  & 0.04 & 5.3     & 0.35    & 0.49       & 1.2   & 0.8    & 2.41       \\ \hline
            VGG-GAN                    & --    & --   & 8.6     & 0.21    & 0.96       & 0.4   & 0.1    & 1.3        \\
            VGG*-GAN                   & 0.36  & 0.15 & 5.2     & 0.24    & 1.29       & 0.7   & 0.1    & 1.24       \\ \hline
            Real data                  & --    & --   & --      & --      & --         & 0.1   & 0.01   & 0.56       \\
        \end{tabular}
    \end{subtable}
    \label{tab:tables}
\end{table}

\subsubsection{User Photorealism Study on CelebA.} We have also performed an informal user study of the photorealism. The study enrolled $30$ subjects unrelated to computer vision and evaluated the photorealism of VGG*-GAN, DFI, CycleGAN and FaceApp on smile and aging/rejuvenation transforms. To assess the photorealism, the subjects were presented quintuplets of photographs unseen during training. In each quintuplet the center photo was an image without the target attribute (e.\,g. real photo of neutral expression), while the other four pictures were manipulated by one of the methods and presented in random order. The subjects were then asked to pick one of the four manipulations that they found most plausible (both in terms of realism and identity preservation). While there was no hard time limit, the users were asked to make the pick as quickly as possible. Each subject was presented overall $30$ quintuplets with $15$ quantuplets allocated for each of the considered attribute. The results in \tab{tables}a show that VGG*-GAN is competitive and in particular considerably better than the other feed-forward method in the comparison (CycleGAN), but FaceApp being the winner overall. This comes with the caveat that the training set of FaceApp is likely to be bigger than CelebA. We also speculate that the diversity of smiles in FaceApp seems to be lower (\fig{baselines-lr}), which is the deficiency that is not reflected in this user study.

\begin{figure}[t]
    \center{\includegraphics[width=\linewidth]{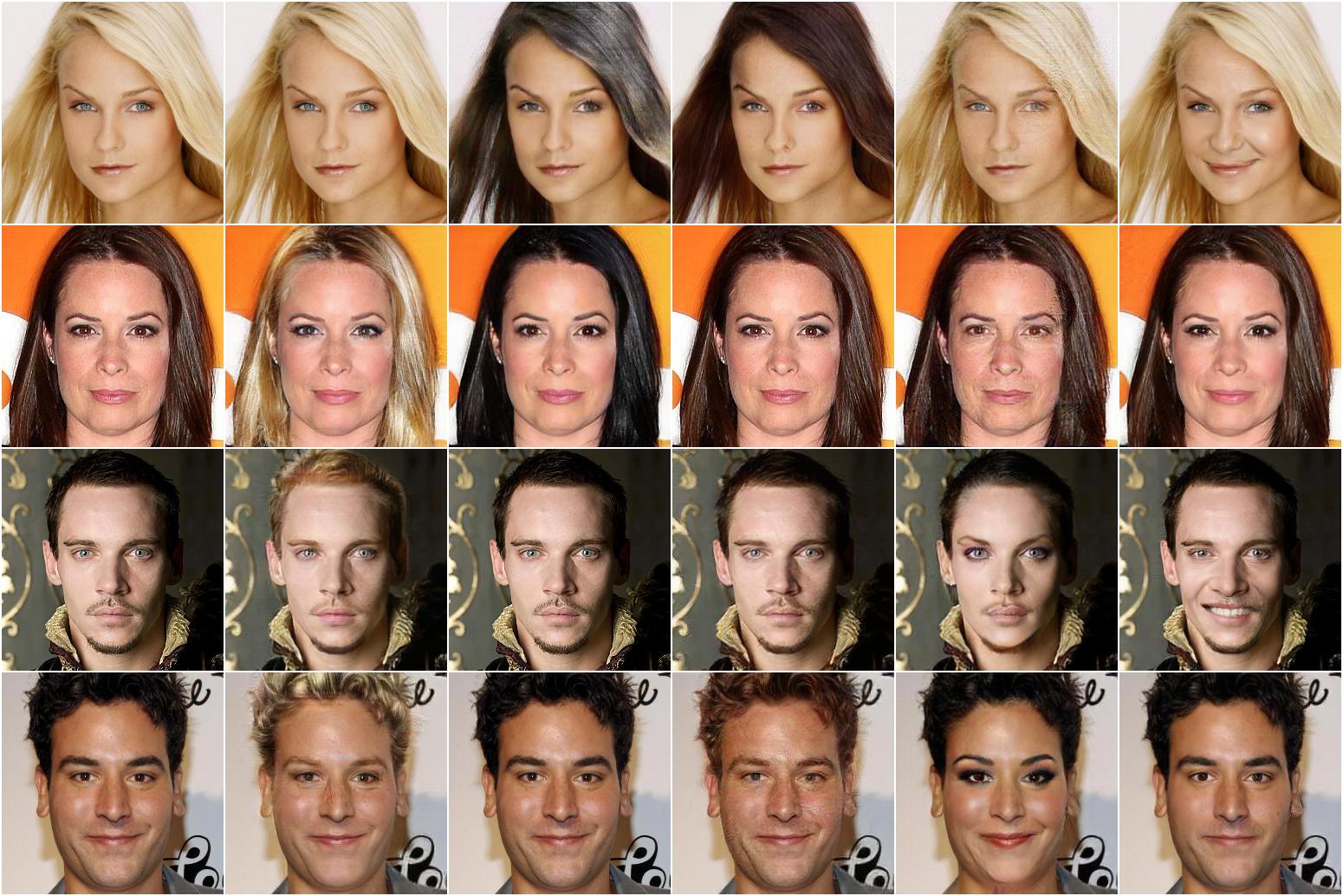}}
    \newcolumntype{A}{>{\centering\arraybackslash}p{0.166\textwidth}}
    \newcolumntype{B}{>{\centering\arraybackslash}p{0.166\textwidth}}
    \newcolumntype{C}{>{\centering\arraybackslash}p{0.166\textwidth}}
    \newcolumntype{D}{>{\centering\arraybackslash}p{0.166\textwidth}}
    \newcolumntype{E}{>{\centering\arraybackslash}p{0.166\textwidth}}
    \newcolumntype{F}{>{\centering\arraybackslash}p{0.166\textwidth}}
    \setlength{\tabcolsep}{0pt}
    \begin{tabularx}{\textwidth}{AAAAAA}
        Input &
        Blond hair &
        Black hair &
        Brown hair &
        Gender swap &
        Smile on/off
    \end{tabularx}
    \caption{Results for VGG*-MS-CycleGAN attribute editing at $256 \times 256$ resolution on Celeba-HQ dataset. Networks have been trained to perform pairwise domain translation between the values of hair color, gender and smile attributes. Digital zoom-in recommended. See~\cite{supmat} for more manipulation examples.}
    \label{fig:faces_best}
\end{figure}

\subsubsection{Quantitative Results on CelebA.} To get objective performance measure, we have used the classifier two-sample test (C2ST)~\cite{Lopez16} to quantitatively compare GANs with the proposed discriminators to other methods. For each method, we have thus learned a separate classifier to discriminate between hold-out set of real images from target distribution and synthesized images, produced by each of the methods. We split both hold-out set and the set of fake images into training and testing parts, fit the classifier to the training set and report the log-loss over the testing set in the Table~\ref{tab:tables}b. The results comply with the qualitative observations: artifacts, produced by DCGAN and DFI are being easily detected by the classifier resulting in a very low log-loss. The proposed system stays on par with a more complex CycleGAN (better on two transforms out of three), proving that a perceptual discriminator can remove the need in two additional networks and cycle losses. Additionally, we evaluated attribute translation performance in a similar fashion to StarGAN~\cite{Choi18}. We have trained a model for attribute classification on CelebA and measured average log-likelihood for the synthetic and real data to belong to the target class. Our method achieved lower log-loss than other methods on two out of three face attributes (see Table~\ref{tab:tables}c).%[INSERT SOMETHING IF VGG*-CYCLEGAN GETS COMPUTED]. %The generators, learned using VGG-based discriminators show the best C2ST scores. 

\subsubsection{Higher Resolution.} We further evaluate our model on CelebA-HQ dataset. Here in order to obtain high quality results we use all proposed regularization methods. We refer to our best model as VGG*-MS-CycleGAN, which corresponds to the usage of VGG* network with dense gradients as an encoder, multi-scale perceptual discriminator based on VGG* network, CycleGAN regularization and pretraining of the generator. Following CycleGAN, we use LSGAN \cite{MaoLXLW16} as an adversarial objective for that model. We trained on $256 \times 256$ version of CelebA-HQ dataset and present attribute manipulation results in \fig{faces_best}. As we can see, our model provides photorealistic samples while capturing differences between the attributes even for smaller amount of training samples (few thousands per domain) and higher resolution compared to our previous tests.

\begin{figure}[t]
    \center{\includegraphics[width=\linewidth]{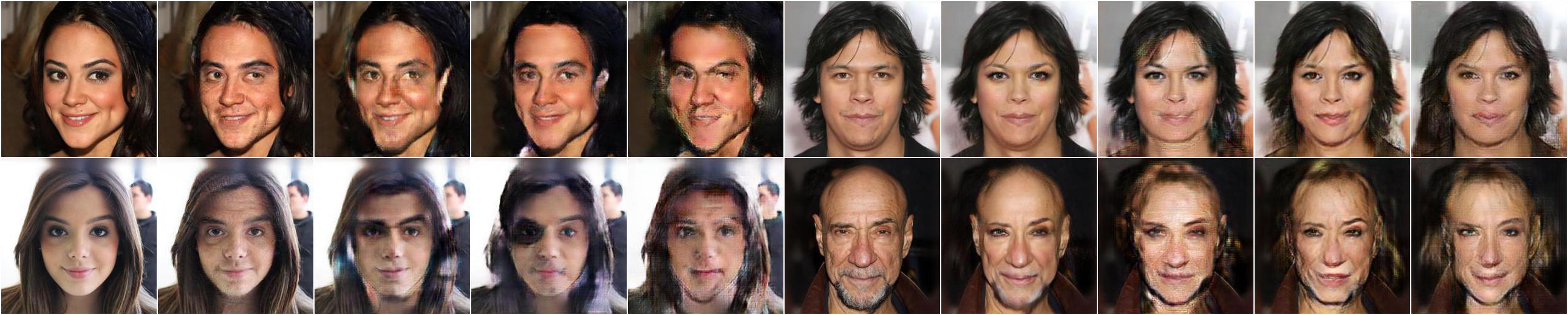}}
    \newcolumntype{A}{>{\centering\arraybackslash}p{0.1\textwidth}}
    \newcolumntype{B}{>{\centering\arraybackslash}p{0.1\textwidth}}
    \newcolumntype{C}{>{\centering\arraybackslash}p{0.1\textwidth}}
    \newcolumntype{D}{>{\centering\arraybackslash}p{0.1\textwidth}}
    \newcolumntype{E}{>{\centering\arraybackslash}p{0.1\textwidth}}
    \newcolumntype{F}{>{\centering\arraybackslash}p{0.1\textwidth}}
    \newcolumntype{G}{>{\centering\arraybackslash}p{0.1\textwidth}}
    \newcolumntype{H}{>{\centering\arraybackslash}p{0.1\textwidth}}
    \newcolumntype{I}{>{\centering\arraybackslash}p{0.1\textwidth}}
    \newcolumntype{J}{>{\centering\arraybackslash}p{0.1\textwidth}}
    \setlength{\tabcolsep}{0pt}
    \begin{tabularx}{\textwidth}{AAAAAAAAAA}
        (a) &
        (b) &
        (c) &
        (d) &
        (e) &
        (a) &
        (b) &
        (c) &
        (d) &
        (e)
    \end{tabularx}
    \caption{We compare different architectures for the discriminator on CelebA-HQ $256 \times 256$ male $\leftrightarrow$ female problem. We train all architectures in CycleGAN manner with LSGAN objective and compare different discriminator architectures. (a) Input, (b) VGG*-MS-CycleGAN: multi-scale perceptual discriminator with pretrained VGG* as a feature network $F$, (c) Rand-MS-CycleGAN:  multi-scale perceptual discriminator with a feature network $F$ having VGG* architecture with randomly-initialized weights, (d) MS-CycleGAN: multi-scale discriminator with the trunk shared across scales (as in our framework), where images serve as a direct input, (e) separate multi-scale discriminators similar to Wang~et~al.~\cite{Wang17}. Digital zoom-in recommended.}
    \label{fig:faces_comp}
\end{figure}

In order to ensure that each of our individual contributions affects the quality of these results, we consider three variations of our discriminator architecture and compare them to the alternative multi-scale discriminator proposed in Wang~et~al.~\cite{Wang17}. While Wang~et~al.~used multiple identical discriminators operating at different scales, we argue that this architecture has redundancy in terms of number of parameters and can be reduced to our architecture by combining these discriminators into a single network with shared trunk and separate multi-scale output branches (as is done in our method). Both variants are included into the comparison  in \fig{faces_comp}. Also we consider \textit{Rand-MS-CycleGAN} baseline that uses random weights in the feature extractor in order to tease apart the contribution of VGG* architecture as a feature network $F$ and the effect of also having its weights pretrained on the success of the adversarial training. While the weights inside the VGG part were not frozen, so that adversarial training process could theoretically evolve good features in the discriminator, we were unable to make this baseline produce reasonable results. For high weight of the identity loss $\lambda_\text{id}$ the resulting generator network produced near-identical results to the inputs, while decreasing $\lambda_\text{id}$ lead to severe generator collapse. We conclude that the architecture alone cannot explain the good performance of perceptual discriminators (which is validated below) and that having pretrained weights in the feature network is important.

\begin{figure}[t]
    \center{\includegraphics[width=\linewidth]{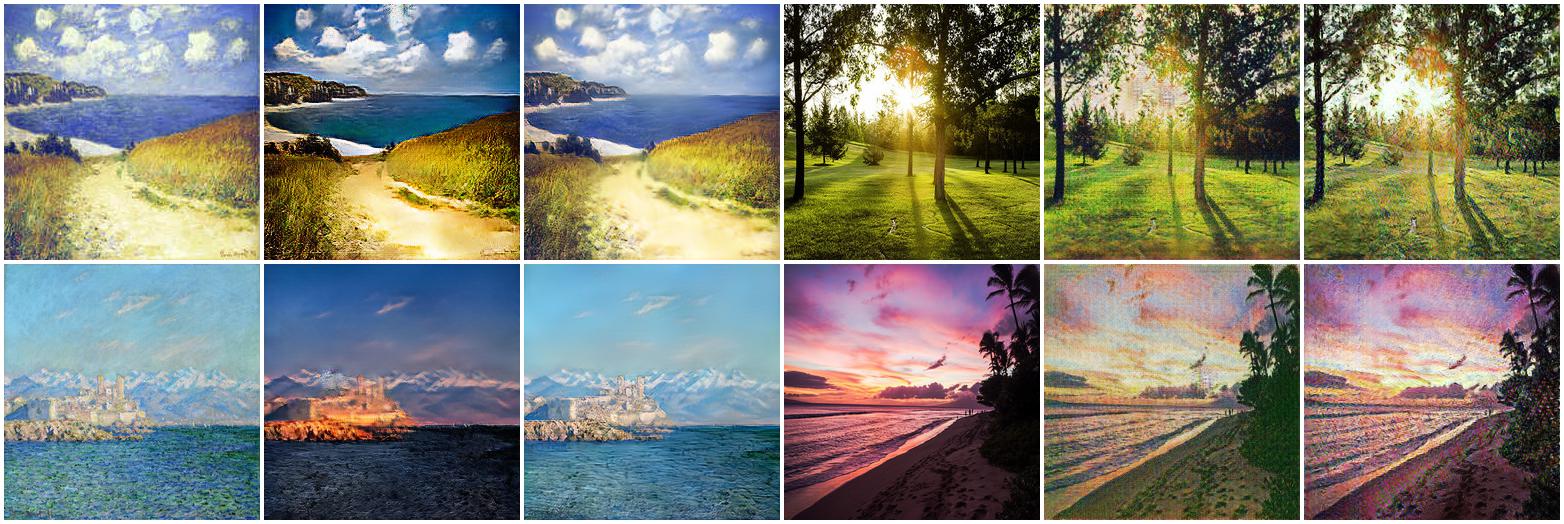}}
    \newcolumntype{A}{>{\centering\arraybackslash}p{0.166\textwidth}}
    \newcolumntype{B}{>{\centering\arraybackslash}p{0.166\textwidth}}
    \newcolumntype{C}{>{\centering\arraybackslash}p{0.166\textwidth}}
    \newcolumntype{D}{>{\centering\arraybackslash}p{0.166\textwidth}}
    \newcolumntype{E}{>{\centering\arraybackslash}p{0.166\textwidth}}
    \setlength{\tabcolsep}{0pt}
    \begin{scriptsize}
    \begin{tabularx}{\textwidth}{AAAAAA}
        Input &
        CycleGAN &
        VGG*-CycleGAN &
        Input &
        CycleGAN &
        VGG*-CycleGAN
    \end{tabularx}
    \end{scriptsize}
    \caption{Comparison between CycleGAN and VGG*-MS-CycleGAN on painting$\leftrightarrow$photo translation task. It demonstrates the applicability of our approach beyond face image manipulation. See \cite{supmat} for more examples.}
    \label{fig:style}
\end{figure}

\begin{figure}[!h]
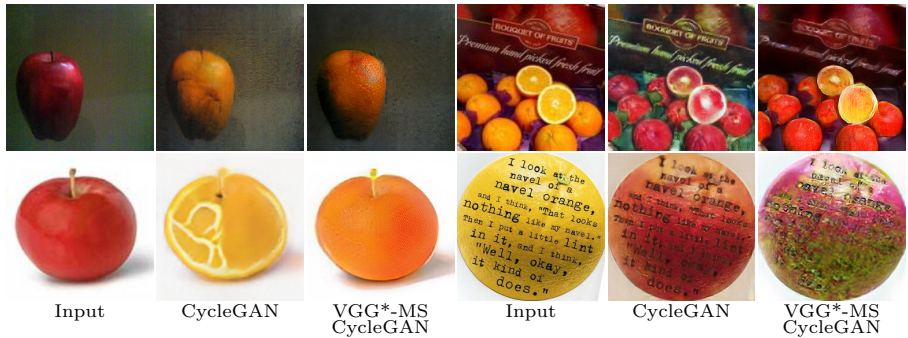

    \def\pth{images/apple2orange/}
    \def\wid{1.95cm}
    \def\linefigure[#1]{\includegraphics[width=\wid]{\pth#1-input-0.png}&
    \includegraphics[width=\wid]{\pth#1-cyclegan-0.jpg}&
    \includegraphics[width=\wid]{\pth#1-ours-0.png}&
    \includegraphics[width=\wid]{\pth#1-input-1.png}&
    \includegraphics[width=\wid]{\pth#1-cyclegan-1.jpg}&
    \includegraphics[width=\wid]{\pth#1-ours-1.png}\\
    }
    \centering
    \begin{scriptsize}
    \setlength{\tabcolsep}{0.7pt}
    \renewcommand{\arraystretch}{0.5}
    \begin{tabular}{@{}cccccc@{}}
    \linefigure[1]
    \linefigure[2]
    Input& CycleGAN& \begin{tabular}[t]{@{}c@{}}VGG*-MS\\CycleGAN\end{tabular}& Input& CycleGAN& \begin{tabular}[t]{@{}c@{}}VGG*-MS\\CycleGAN\end{tabular}
    \end{tabular}
    \end{scriptsize}
    \caption{Apple$\leftrightarrow$orange translation samples with CycleGAN and VGG*-MS-CycleGAN are shown. Zoom-in recommended. See \cite{supmat} for more examples.}
    \label{fig:apple2orange}
\end{figure}

\subsubsection{Non-face Datasets.} While the focus of our evaluation was on face attribute modification tasks, our contribution applies to other translation tasks, as we verify in this section by performing qualitative comparison with the CycleGAN and VGG*-MS-CycleGAN architectures on two non-face domains on which CycleGAN was originally evaluated: an artistic style transfer task (Monet-photographs) in \fig{style} and an apple-orange conversion in \fig{apple2orange} (the figures show representative results). To achieve fair comparison, we use the same amount of residual blocks and channels in the generator and the same number of downsampling layers and initial amount of channels in discriminator both in our model and in the original CycleGAN. We used the authors' implementation of CycleGAN with default parameters. While the results on the style transfer task are inconclusive, for the harder apple-to-orange task we generally observe the performance of perceptual discriminators to be better.

\subsubsection{Other Learning Formulations.} Above, we have provided the evaluation of the perceptual discriminator idea to unaligned image translation tasks. In principle, perceptual discriminators can be used for other tasks, e.g.\ for unconditional generation and aligned image translation. In our preliminary experiments, we however were not able to achieve improvement over properly tuned baselines. In particular, for aligned image translation (including image superresolution) an additive combination of standard discriminator architectures and perceptual losses performs just as well as our method. This is not surprising, since the presence of alignment means that perceptual losses can be computed straight-forwardly, while they also stabilize the GAN learning in this case. For unconditional image generation, a naive application of our idea leads to discriminators that quickly overpower generators in the initial stages of the game leading to learning collapse. 
\section{Summary} \label{section_conclusion}

We have presented a new discriminator architecture for adversarial training that incorporates perceptual loss ideas with adversarial training. We have demonstrated its usefulness for unaligned image translation tasks, where the direct application of perceptual losses is infeasible. Our approach can be regarded as an instance of a more general idea of using transfer learning, so that easier discriminative learning formulations can be used to stabilize and improve GANs and other generative learning formulations.
%We have shown that our model performs on-par with state-of-the-art image attribute translation frameworks for unpaired domain adaptation problems. We have also shown that our perceptual discriminator architecture is scalable in terms of input image resolution and can be combined with existing approaches to improve the results. %We are positive that this approach can be applied to a higher resolution data given big enough dataset and enough training time, but in our paper we've already shown that our approach works for higher resolution and with better quality compared to other feed forward architectures, setting new state-of-the-art in unpaired image translation problems.

\subsubsection*{Acknowledgements.}
This work has been supported by the Ministry of Education and Science of the Russian Federation (grant 14.756.31.0001).

% \FloatBarrier
% \ifnum\value{page}>14 \errmessage{Number of pages exceeded!!!!}\fi

\clearpage
\bibliographystyle{splncs04}
\bibliography{2992.bbl}   % name your BibTeX data base

\end{document}